# Convolutional Neural Networks for Sleep Stage Scoring on a Two-Channel EEG Signal

Enrique Fernandez-Blanco[1], Daniel Rivero[1], Alejandro Pazos[1,2]

enrique.fernandez@udc.es
daniel.rivero@udc.es
alejandro.pazos@udc.es

[1]Faculty of Computer Science. CITIC. University of A Coruna. A Coruña, 15071, Spain
[2]Instituto de Investigacion Biomedica de A Coruña (INIBIC), Complexo Hospitalario Universitario de A Coruña (CHUAC), A Coruña, 15006, Spain

* Corresponding Author: enrique.fernandez@udc.es Phone:+34881016014

*Hightlights*

- Sleep Stage scoring with automatic feature extraction
- Downsized solution with respect to state-of-the-art
- Applicability of Convolutional Neural Networks for simultaneous signal processing
- Statistical analysis of the advantages of using two channels instead of other alternatives, such as using only one

*Abstract.-*
Sleeping problems have become one of the major diseases all over the world. To tackle this issue, the basic tool used by specialists is the Polysomnogram, which is a collection of different signals recorded during sleep. After its recording, the specialists have to score the different signals according to one of the standard guidelines. This process is carried out manually, which can be highly time consuming and very prone to annotation errors. Therefore, over the years, many approaches have been explored in an attempt to support the specialists in this task. In this paper, an approach based on convolutional neural networks is presented, where an in-depth comparison is performed in order to determine the convenience of using more than one signal simultaneously as input. Additionally, the models were also used as parts of an ensemble model to check whether any useful information can be extracted from signal processing a single signal at a time which the dual-signal model cannot identify.
Tests have been performed by using a well-known dataset called expanded sleep-EDF, which is the most commonly used dataset as benchmark for this problem. The tests were carried out with a leave-one-out cross-validation over the patients, which ensures that there is no possible contamination between training and testing. The resulting proposal is a network smaller than previously published ones, but which overcomes the results of any previous models on the same dataset. The best result shows an accuracy of 92.67% and a Cohen's Kappa value over 0.84 compared to human experts.

# 1. Introduction

Among the essential body functions like breathing, eating or drinking, sleeping is probably the most problematic one nowadays. According to the US government through its Centers for Control of Disease and Prevention (CDC), about 9 million citizens have frequent problems to develop good quality sleep and end up resorting to sleeping pills (Ford et al. 2014). In parallel, recent studies (Stranges et al. 2012; Chong et al. 2013) have estimated that at least 15% of adult population might have some kind of sleeping problem or poor-quality sleep as a result of a number of issues. Moreover, the World Health Organization (WHO) (2015) claimed that a good quality sleep was one of the most important factors for good health while sleeping problems were directly related to other diseases, including depression, stress or early cardiac diseases.

As a consequence, new units focused on the study and treatment of sleeping problems have been created in hospitals all over the world. The physicians in these units have as their main tool for their work the records obtained during their patients' sleep. These records, called polysomnography (PSG), may include a great variety of signals such as Electrocardiograms, Electroencephalograms, respiratory signals or movement records. Among these signals, the most important one is the Electroencephalogram (EEG) because it is the most reliable to determine the sleep stage a patient is in.

The interpretation of an EEG is a highly time-consuming activity (Akben and Alkan 2016), which usually requires a specialist and it is deeply dependent on the expert's expertise. From EEGs, physicians mainly split the records in two main stages: Rapid-Eye Movement (REM) and non-Rapid-Eye Movement (NREM). Then, the latter stage is also further subdivided in different substages. This process can follow two different guidelines proposed by Rechtschaffen & Kales (R&K) (Rechtschaffen and Kales 1968) and the American Academy of Sleep Medicine (AASM) (Iber et al. 2007), respectively. Therefore, the described annotating process tends to lead to a misclassification as a result of fatigue (Boashash and Ouelha 2016) or the high amount of data. In fact, the dependency on the expert has been measured as an agreement around 80% for interobserver (Norman et al. 2000) and more or less the same amount for intraobserver (Wendt et al. 2015).

In this work, an automatic sleep stage scoring system based on Convolutional Neural Networks (CNNs) is proposed to perform sleep scoring while reducing the dependency on experts to identify the features. Each layer of a CNN is focused on refining the features of the previous one by training the applied filters. Therefore, this kind of neural network is able to train whichever features are more suitable to solve a problem without requiring an expert to identify them. By using this simple principle, the CNN has been successful applied in dealing with many problems such as image classification (Krizhevsky et al. 2017), synthetic image creation (Goodfellow et al. 2016), Natural Language Processing (Deng and Yu 2014), etc.

However, nowadays it is still unclear whether this kind of artificial neural network is able to find the relationships between the simultaneously recorded signals in order to improve the solution to the problem.

This paper is structured as follows: an introduction; in section 2, there is a summary and discussion of the main references; section 3 contains an explanation of the data used in this work and the method proposed to solve the problem raised; section 4 and 5 contain the description of the test and the discussion of the results comparing them with previously published works; section 6 contains the conclusions of the work, whereas section 7 exposes some future lines of work that might be conducted from the results described in sections 4 and 5.

## 2. Background

Electroencephalogram (EEG) is one of the very few methods to study the brain and its behaviour. One of the biggest issues of this method is the fact that the manual labelling has to be made by specialists, resulting in a monotonous task which is particularly prone to fatigue errors. Therefore, there is a long list of works that focused on tackling the issue of automatic labelling of EEGs for different problems and diseases, for example, to identify epileptic seizures using different methods (Tzallas et al. 2007; Übeyli 2010; Fernández-Blanco et al. 2013; Hassan and Subasi 2016).

Another common problem in hospitals, which has not attracted as much attention as others, like epilepsy, is the labelling of polysomnography (PSG). These tests are composed of several complex signals such as EEG, electrocardiogram or respiratory records, which are simultaneously recorded during a night in a sleeping unit at a hospital.

One of the first attempts to automate the scoring of the EEGs contained in a PSG was the study conducted by (Berthomier et al. 2007). The authors used Fuzzy Logic in combination with an iterative method to label the original sleep-EDF dataset (Kemp et al. 2000) to classify the records of 8 patients, half of whom were using drugs to sleep. The same dataset was also used in (Hsu et al. 2013), where the authors used a combination of the energy from different frequency bands and a neural network to perform the classification.

Other authors have focused their attention on the temporal line to extract the features which made the classification possible. For example, (Sheng-Fu Liang et al. 2011) proposed a decision tree to classify the features, and (Doroshenkov et al. 2007) used a Hidden Markov Model to perform the classification. Alternatively, other authors have explored other approaches for automatic feature extraction, such as (Liang et al. 2012), where a combination of a multiscale entropy with a simple linear discriminant analysis (LDA) (McLachlan 1992) was used to score the sleeping records. Other works have preferred to focus their attention on other features with a high variety of classification methods, such as statistical features with bagging (Hassan et al. 2015a), power spectral density with artificial neural networks (Ronzhina et al. 2012), graph theory features with support vector machines (Zhu et al. 2014) or moment features with boosting (Hassan et al. 2015b).
On the other hand, some proposals – instead of keeping the timeline as it is – have transformed the search space in order to improve the classification. For example, (Vural and Yildiz 2008) applied a Karhunen–Love transform to extract hybrid features for the classification, and (Hassan and Subasi 2017) used a wavelet transform to extract the features and execute a boosting classifier.

Finally, the works which use Deep Learning (LeCun et al. 2015) as an approach ought to be pointed out because they are the most closely related to the method discussed in this paper. Deep Learning techniques, such as Random Belief Networks (Hinton 2009) and Convolutional Neural Networks (Lecun et al. 1998) are only two well-known examples of techniques framed under the term Deep Learning. The principle of those networks is simple: each new layer of a neural network extracts higher level features from the information on the input. Applying this principle to EEGs, the first attempt to score sleeping stages can be found in (Tsinalis et al. 2016), whose authors proposed to use the autoencoders to solve the labelling problem in a 20-patient dataset known under the name of sleep-EDF-expanded described in (Kemp et al. 2000). Moreover, using the sleep-EDF-expanded in combination with another dataset known as MASS (O'Reilly et al. 2014), (Supratak et al. 2017) proposed a model called Deep Sleep which is based on a two-pipeline network. Finally, (Sors et al. 2018) proposed an approach based on the feature extraction capabilities of the Convolutional Neural Networks (CNN) (Krizhevsky et al. 2017), which was applied to a dataset called SNNS-1 (Quan et al. 1997). The latter work was the only one to use this dataset, making the comparison of results particularly difficult.

## 3. Materials and Methods

### 3.1. Data Description

The dataset described in (Kemp et al. 2000), known as sleep-EDF-extended, which was obtained from Physionet (Goldberger et al. 2000), is probably the most frequently used when attempting to address the sleeping scoring issue. It was also used in this work to perform the experiments.

The dataset contained 61 Polysomnograms (PSGs) recorded from two experiments. In the first, 20 healthy patients were recorded in two consecutive days during approximately 20 hours each. In the second experiment, records were obtained from people under medication due to sleeping problems for a period between 6 to 8 hours.

Although the second group is highly interesting, the only data available were from the first night, when the patients were without medication. Therefore, that limits the usability of that section and the spotlight of this work was on the first group, in order to maintain the same conditions as in other previously published works, allowing for a fuller and fairer comparison. Consequently, only the healthy patients were included and, for each patient, two recorded PSGs were used, except for three patients for whom the data corresponding to only one night were available, as explained in the original paper.

Recorded PSGs contained, among other data, two EEG signals from Fpz-Cz and Pz-Oz electrode locations (Fig. 1) which were recorded at a sampling frequency of 100 Hz. Another important signal in the record is the hypnogram, which contained the labels assigned by a specialist according to the R&K guidelines (Rechtschaffen and Kales 1968). Therefore, one of the following labels was assigned to each piece of 30s: Awake, REM, Stage 1, Stage 2, Stage 3, Stage 4, Movement time and Unknown.

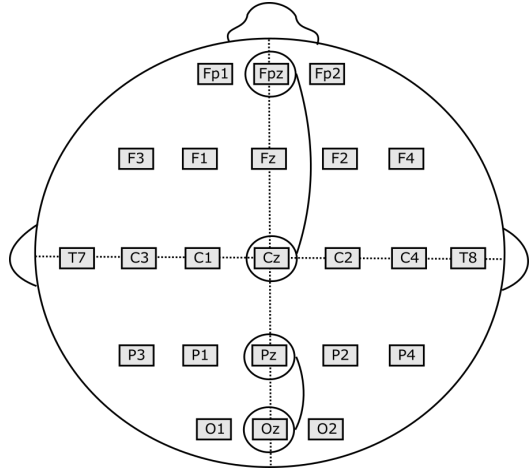

Fig. 1.- Diagram of the possible location for the connector of an EEG

These data were adapted according to the new AASM standard (Iber et al. 2007). Consequently, 'unknown' and 'movement time' sections were discarded. Stages 3 and 4 were joined in a single class corresponding to light sleep and Awake, REM, Stage 1 and Stage 2 were kept. A summary of those transformations can be seen in Table 1. According to the aforementioned adaptations, the dataset contained 110,925 sections of 30s distributed as shown in Table 1.

Table 1: Number of 30s-length sections for each annotation class according to the R&K and AASM guidelines

| R&K State | AWA | REM | S1 | S2 | S3 & S4 | Unknown & Movement |
|---|---|---|---|---|---|---|
| AASM State | Awaked | REM | N1 | N2 | N3 | - |
| Num. of sections | 68675 | 2662 | 16791 | 5501 | 7296 | - |

The main reason behind choosing this dataset and the described adaptation is to increase the number of works to compare it with, because several works have previously used the same approach.

## 3.2. Convolutional Neural Networks

Based on the early works of Fukushima (Fukushima 1980) and Lecun (Lecun et al. 1998), convolutional neural network (CNN) and, in general, Deep Learning have meant an important step forward in many knowledge areas by becoming the state-of-the-art for many problems. These works established a hierarchy of layers, where each neuron receives as input a spatially-close related piece of information. Each neuron on a layer receives different input data from a sliding window over the signal or image. The weights are the same for all neurons, unlike classical neural networks, where they are different. Consequently, the result is the convolution of an input feature map $X^{(l-1)}$ with a set of learnable filters $W^{(l)}$, adding biases $b^{(l)}$ and, finally, applying some kind of transmission function $g$, as described in Eq. 1

$$X^{(l)} = g^l(X^{(l-1)} * W^{(l)} + b^{(l)}) \qquad (1)$$

If this scheme is repeated several times, the result is a network where each layer extracts a more general information from the information on the previous layer, but conditioned by the spatial relationship (LeCun et al. 2015). Therefore, CNNs are usually composed of a number of convolutional layers which extract the features of the signals or images. These extracted features are then followed by some kind of classification technique such as a fully-connected perceptron or a softmax regression layer, which gives the outputs class probabilities according to the features extracted by the convolutional part.

This scheme has been successfully used with a wide range of applications, although most of these works are mainly related with image processing, such as face recognition (Taigman et al. 2014), and image classification (Russakovsky et al. 2015), while signal processing applications are quite limited to natural language processing (Dahl et al. 2012).

### 3.3. Proposed architecture

This paper proposes an architecture for scoring the sleep stages of PSG by using a convolutional neural network (CNN) much simpler than other proposed in recent works on similar problems (Supratak et al. 2017; Sors et al. 2018).

These aforementioned works used only one of the available signals in the dataset whereas, for example, in the one described by (Kemp et al. 2000), there are always 2 simultaneous recorded EEGs available. The architecture proposed in this work was tested separately with each signal, and with both signals together as inputs, resulting in three different systems. As mentioned in the description of the dataset, signals were labelled by a physician each 30 s, therefore the size of the input would be exactly the sampling rate by the number of seconds, i.e. 3000 inputs or 6000 depending on the number of signals used as inputs.

After that, a series of 7 convolutional layers with kernel sizes running down from 7 to 3 were set, in an attempt to extract more general features at the beginning, while more specific and complex features were extracted in the final stages. The initial and final sizes were chosen according to several brain machine works (Sakhavi et al. 2018), which set these sizes as the most suitable for the extraction of EEG features in time domain.

Therefore, 20 features for each application of the kernel were the output of each convolutional layer. That output was later modified by a ReLu transmission function. The output of each convolutional layer passed through a pooling layer which performed a maximum operation on each two elements of the convolutional output. These pooling layers shifted the attention of the network to the peaks of the signal and later to the most promising features throughout time. On the other hand, the depth of the network was the result of a bunch of preliminary tests. As shown in Fig. 2, the results of these convolutions are a set of 400 high-level features which were fed in a fully connected layer. In that figure, to simplify the representation, combinations of convolution and max pooling layers were represented as a single functional block with 3 parameters, where the first one (K) represented the length of the filter to be applied, the second one (F) was the number of features that were going to be extracted and the last one (M) was the size of the kernel for the max pooling operation. The output layer would be composed of 5 neurons, one for each class, which used a softmax function to determine the belonging to each possible class in the shape of a one-hot-encoding array. Additionally, in order to improve the generalization, a Dropout layer (Srivastava et al. 2014) with a value of 0.5 was previously applied to this last layer in order to prevent overfitting during training.

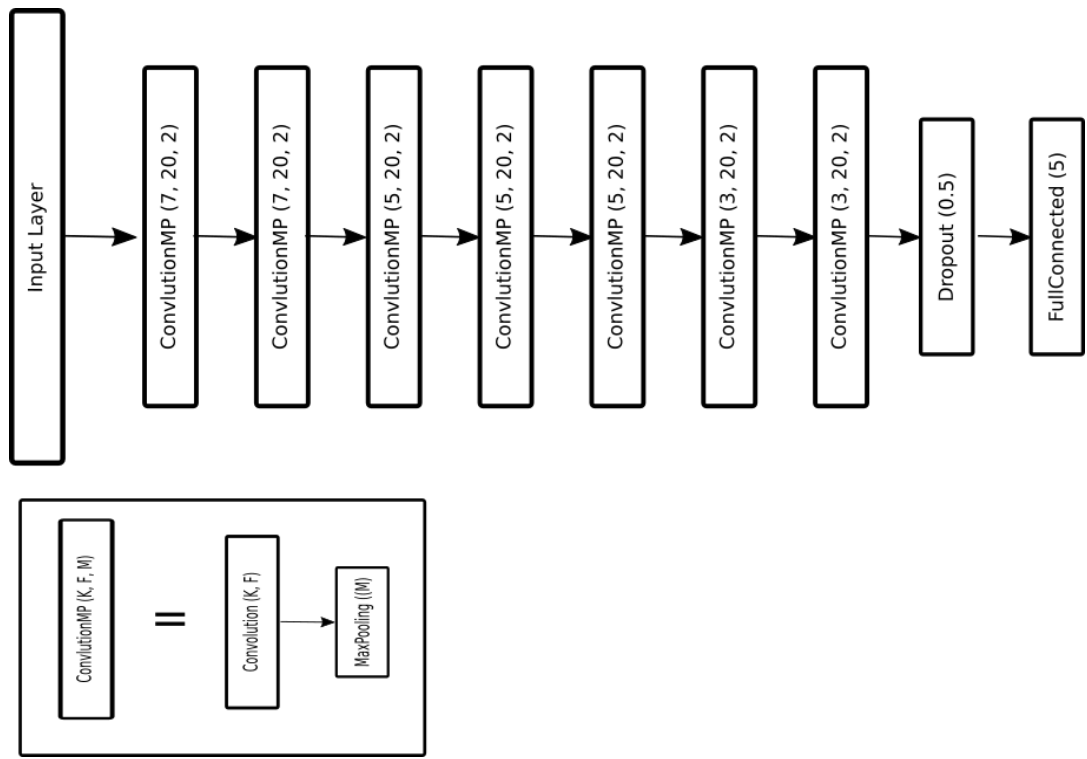

Fig. 2: Diagram of the proposed architecture. The numbers in brackets correspond to the length of the filter and the number of filters.

### 3.4. Training Process

The training process of the architecture, described in section 3.3, was performed by following a variant of the K-fold cross-validation (Mosteller and Tukey 1965) known as leave-one-out. According to the general description, the K-fold cross-validation divides the data set in K subsets and performs K experiments. In each of these experiments, a different subset was used for testing, while the remaining K-1 subsets were used for training. This process was repeated K times and, finally, the average of the K results in tests was calculated. If this process is taken to the limit and the number K is equal to the number of elements in the dataset, what we get is the aforementioned variant called leave-one-out.

In this particular case, in order to check the generalization of the results, the scheme was applied to the number of patients described in section 3.1. The training was therefore repeated 20 times, leaving out a single patient with all his signals in order not to corrupt the training process and keep it as close to a real application as possible. Thus, 20 different experiments were carried out for each one of the three possible inputs, i.e., only Fp-Cz signal, only Pz-Oz signal or both signals simultaneously. In each of these training processes, the training set contained 19 patients and the signal or signals from the other patient was/were used as test set.

Once the test dataset was separated from the training dataset, 10% of the remaining data were also reserved for validation during training. The training process was then carried out by following a mini-batch scheme with size 20. Those mini-batches were fed to the networks before updating the weights according to a gradient descent algorithm. The validation dataset was evaluated each time the remaining data was fed. This process was stopped after reaching a threshold of 100 epochs or after not being able to improve the validation loss after 10 epochs.

As a loss function for each one of those 20 trainings, categorical cross entropy was used. This function measures the similarity between the distribution of the model output with respect to the ground truth distribution when the outputs are in the shape of a probability like the softmax layer does. In Eq. 2, the formula can be seen, where **p** represents the distribution of the ground truth, **q** represents the distribution of the model results, and **i** represents each particular response evaluated.

$$H(p, q) = - \sum_i p(i) \, log(q(i)) \qquad (2)$$

This loss function has been lately optimised by using the Adaptive Moment Estimation algorithm (Kingma and Ba 2015) with an initial value of 0.001, while using batches of size 20 segments at the same time.

### 3.5. Ensemble - Majority voting

In addition to the proposed architecture, an ensemble model was also developed. Ensemble models is a topic that over the years has attracted much interest in the machine learning community. These models are meta-classifiers which combine several individual classifiers in order to obtain a more robust method. The ensemble of models can be carried out by many different methods (Kuncheva 2002), such as majority voting, weighted majority voting and naive Bayes combination.

In this work, in addition to the proposed architecture explained in section 3.3 with 1 or 2 signals as inputs, an ensemble model was also tested by combining each of the best networks obtained in each fold in the three different approaches: using only the Fpz-Cz signal, using only the Pz-Oz signal and with both of them. To perform the combination, the majority voting system (Kuncheva and Alpaydin 2007) was used due to its simplicity and little computational overhead.

When applying this method, each model performs the classification separately and, then, the meta-classifier chooses the most commonly used class among the members of the ensemble. In the case of a tie, an unknown label is given. Therefore, if $M_k$, $k \in \{1, 2, ..., N\}$ represents each member of an ensemble of $N$ classifiers, where each classifier provides a label $l$, $l \in \{1, 2, ... c\}$ as $y_k$ output. Then, the majority vote (MV) could be defined as in Eq. 3.

$$MV = argmax_l \sum_{i=1}^{N} \delta(l, y_i) \qquad (3)$$

where $\delta(a, b) = 1$ if $a = b$ and $\delta(a, b) = 0$ if $a \neq b$. As already explained, in case of a tie between two different classes, an unknown label is returned and that window is not classified.

## 3.6. Performance measures

In order to get an idea of the goodness of a particular model on the test dataset, there are three main measures usually provided by the different works as performance indicators: accuracy, Cohen's kappa and F1-score.

Generally speaking, the confusion matrix is composed of the number of True Positive (TP), False Positive (FP), False Negative (FN) and True Negative (TN) cases. The accuracy measures the number of correctly classified examples, and is calculated by means of Eq. 4.

$$Accuracy = \frac{TP+TN}{TP+FP+FN+TN} \quad (4)$$

While, F1 score represents how representative are the TP cases identified by the model and how many representative cases were identified. This score can be calculated by means of a harmonic mean between the precision and the recall of the model (Eq. 5).

$$F1 - score = \frac{2TP}{2TP+FN+FP} \quad (5)$$

Finally, Cohen's Kappa is a score used to measure how alike are two classifiers once the agreement by chance is removed. The score can be used, for example, to compare how similar are a developed model and a human expert. To calculate it, the agreement probability ($P_e$) and the observed agreement proportion ($P_o$) were needed before applying Eq. 6

$$\kappa = \frac{P_o - P_e}{1 - P_e} \quad (6)$$

# 4. Results

The main idea behind the test presented in this section is to evaluate whether there is a difference in the performance when the EEGs from a PSG are automatically scored by the same network, but changing the input between 3 possibilities: Fz-Cz signal, Pz-Oz signal, and both of them. Moreover, an ensemble model of the best network for each run was also tested aimed at determining whether there is any valuable information that the single-input-signals model can identify whereas the dual-input-signal model cannot.

The problem to be solved is to assign one of the 5 possible labels defined in the AASM guideline, i.e., Awake, REM, N1, N2, N3, to each section contained in the dataset without any additional filtering.

As previously mentioned in section 3.4, tests were conducted by applying the leave-one-out method to the sleep-EDF dataset described in section 3.1. Consequently, the dataset was divided in 20 subsets corresponding to the 20 patients in the dataset. Therefore, each patient was used only once for testing purposes and 19 times for training, thus making up the 20 repetitions in the entire process.

Therefore, four experiments were conducted: first, using only the Fpz-Cz channel as an input; second, using the Pz-Oz channel as an input; third, using both signals as inputs, and, finally, an ensemble of the three best networks for each fold.

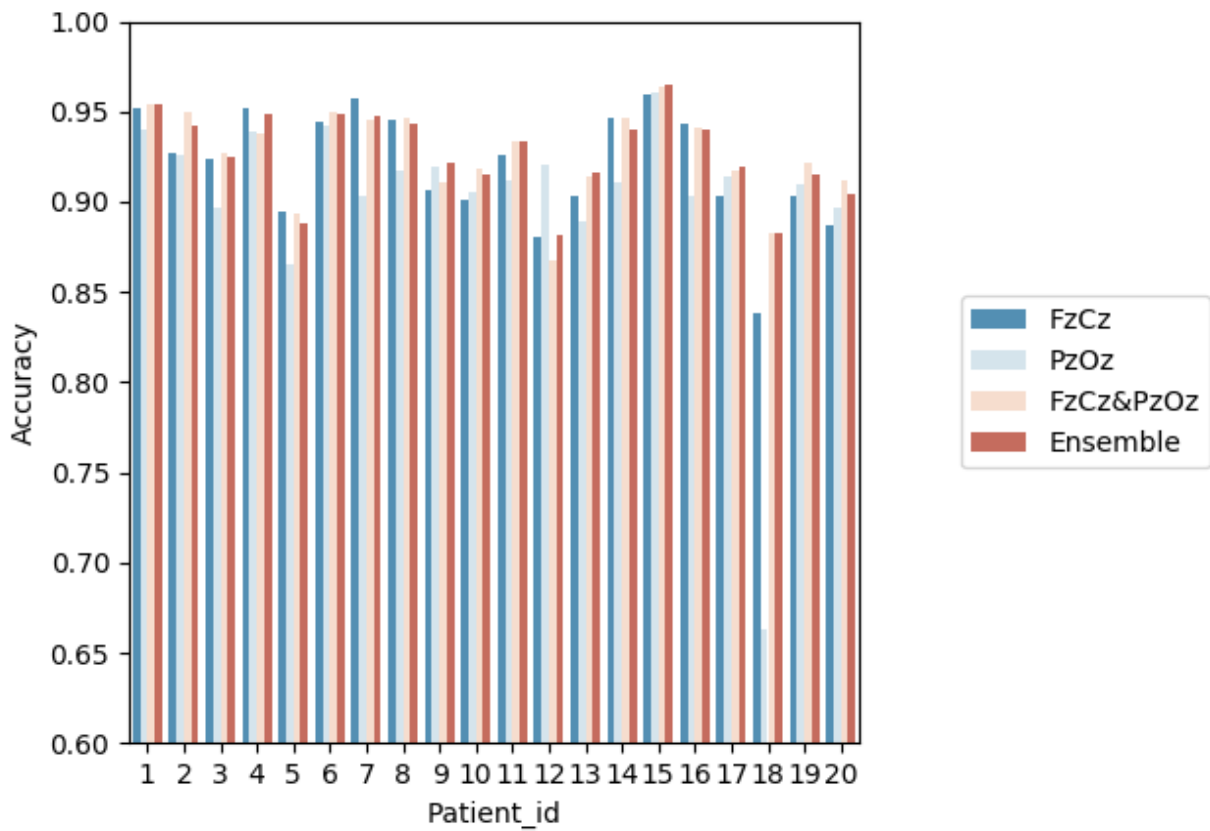

Fig. 3. - Accuracy results for each patient

The entire process was implemented in Python using Keras as framework in combination with Tensorflow as underlying support and it took around 20 hours to execute it on an Intel i7 computer with 16 Gb of RAM and an NVidia Titan X graphic card.

Figures 3 and 4 show the results for each patient in accuracy and Cohen's Kappa, respectively. Both figures show the four different approaches performed in this work: using channel FzCz, using channel PzOz, both channels at the same time, and an ensemble of the best models. A closer look to the results in the Figure 3 pointed out two cases. First, patient number 12 was significantly better classified by the model using only PzOz channels. On the other hand, patient number 18 presented a clearly worse behaviour of the models with a single channel as an input whereas the ensemble and the two-signal input performed clearly better. Both cases confirm this behaviour if the attention is focused on the kappa value represented in Figure 4, which could indicate that these two patients were wrongly scored by the specialist or they were particularly difficult cases, with little in common with the remaining patients.

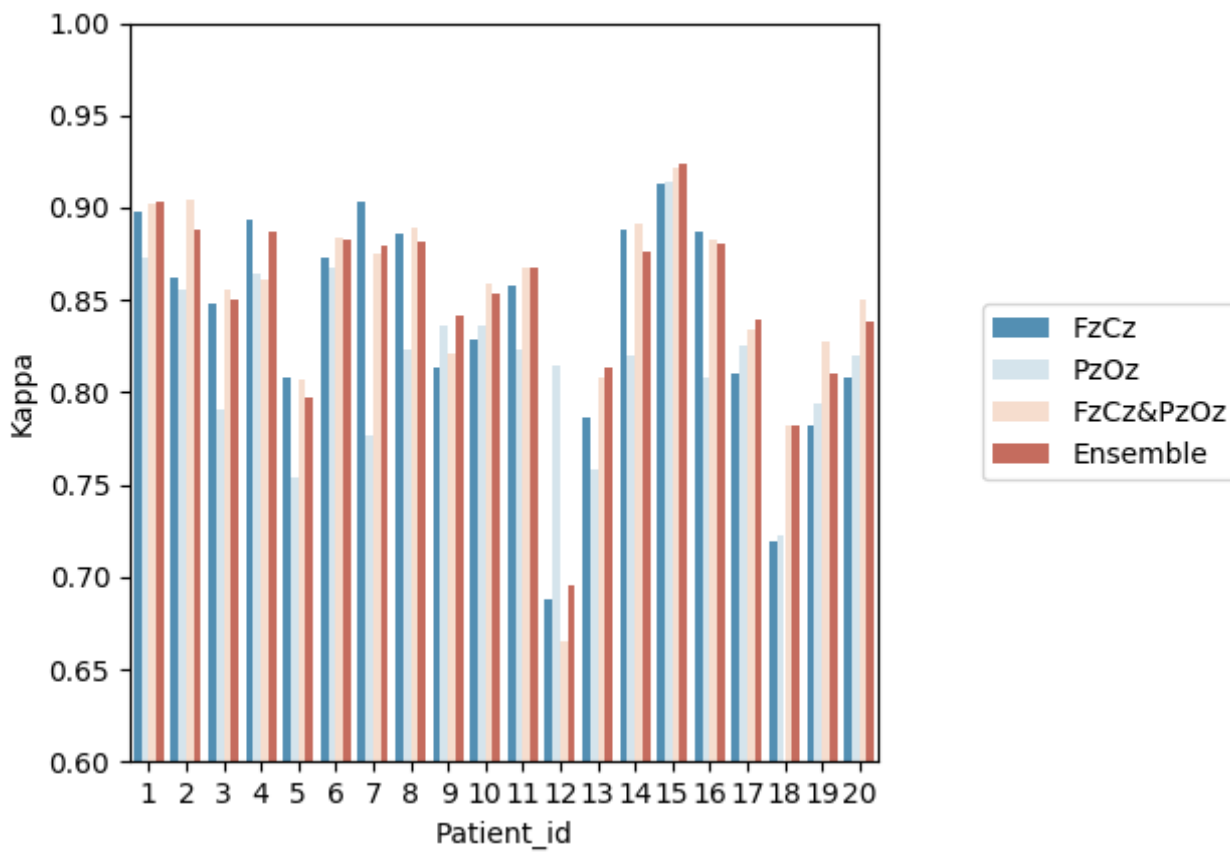

Fig. 4. - Cohen's Kappa results for each patient

Further details on these results are given in Table 2, which shows the accuracy, the Cohen's Kappa with respect to the human expert classified, and the F1-Score. In addition, the last row shows the average of the values and the standard deviation among the patients.

With these results, a Kolmogorov-Smirnov test (Smirnov 1948) was carried out to determine the normality of the results distribution for each experiment, resulting in the rejection of the null hypothesis. Consequently, under the non-normality hypothesis – having tested the resulting model against each patient – a Wilcoxon test (Wilcoxon 1945) for paired samples was performed under a significant level of 0.05. When the double-input-signal model was compared with the single-signal-input models, the statistical test showed a significant difference. More specifically, the test returned 0.0251 when the double signal was compared with the model using only the Fpz-Cz channel, and 0.0028 when the Pz-Oz channel is used as an input. On the other hand, the comparison between the double-signal-input model and the ensemble model did not show any difference.

Table 2: Accuracy, Cohen's Kappa and F1 score for each experiment and test fold.

| Test Patient ID | Fpz-Cz | | | Pz-Oz | | | Fpz-Cz & Pz-Oz | | | Ensemble | | |
|---|---|---|---|---|---|---|---|---|---|---|---|---|
| | Acc | K | F1 | Acc | K | F1 | Acc | K | F1 | Acc | K | F1 |
| 01 | 0.9522 | 0.8981 | 0.7665 | 0.9401 | 0.8728 | 0.7275 | 0.9538 | 0.9027 | 0.7728 | 0.9544 | 0.9036 | 0.6348 |
| 02 | 0.9269 | 0.8625 | 0.7515 | 0.9257 | 0.8561 | 0.7101 | 0.9501 | 0.9039 | 0.8106 | 0.9419 | 0.8885 | 0.6362 |
| 03 | 0.9238 | 0.8484 | 0.7575 | 0.8965 | 0.7902 | 0.6850 | 0.9270 | 0.8552 | 0.7664 | 0.9249 | 0.8500 | 0.6261 |
| 04 | 0.9515 | 0.8940 | 0.7832 | 0.9389 | 0.8640 | 0.6995 | 0.9379 | 0.8608 | 0.7353 | 0.9490 | 0.8870 | 0.6230 |
| 05 | 0.8943 | 0.8083 | 0.7366 | 0.8650 | 0.7538 | 0.6170 | 0.8934 | 0.8073 | 0.7196 | 0.8882 | 0.7976 | 0.5863 |
| 06 | 0.9444 | 0.8725 | 0.7250 | 0.9424 | 0.8677 | 0.7151 | 0.9495 | 0.8841 | 0.7584 | 0.9488 | 0.8825 | 0.6072 |
| 07 | 0.9575 | 0.9028 | 0.7886 | 0.9037 | 0.7763 | 0.6338 | 0.9453 | 0.8748 | 0.7729 | 0.9473 | 0.8793 | 0.6345 |
| 08 | 0.9450 | 0.8857 | 0.8042 | 0.9176 | 0.8235 | 0.6808 | 0.9462 | 0.8887 | 0.7764 | 0.9430 | 0.8815 | 0.6379 |
| 09 | 0.9063 | 0.8136 | 0.6588 | 0.9190 | 0.8360 | 0.6797 | 0.9112 | 0.8209 | 0.6621 | 0.9214 | 0.8415 | 0.5575 |
| 10 | 0.9011 | 0.8284 | 0.7224 | 0.9053 | 0.8362 | 0.7185 | 0.9184 | 0.8587 | 0.7365 | 0.9150 | 0.8534 | 0.6173 |
| 11 | 0.9265 | 0.8578 | 0.7418 | 0.9116 | 0.8227 | 0.6119 | 0.9338 | 0.8679 | 0.7289 | 0.9338 | 0.8679 | 0.5932 |
| 12 | 0.8805 | 0.6874 | 0.6460 | 0.9205 | 0.8148 | 0.6734 | 0.8677 | 0.6651 | 0.6091 | 0.8818 | 0.6959 | 0.5365 |
| 13 | 0.9033 | 0.7864 | 0.7137 | 0.8891 | 0.7577 | 0.6562 | 0.9137 | 0.8085 | 0.7301 | 0.9167 | 0.8139 | 0.6012 |
| 14 | 0.9460 | 0.8885 | 0.8016 | 0.9111 | 0.8196 | 0.6708 | 0.9470 | 0.8914 | 0.8199 | 0.9396 | 0.8760 | 0.6478 |
| 15 | 0.9595 | 0.9128 | 0.7730 | 0.9602 | 0.9140 | 0.7561 | 0.9636 | 0.9221 | 0.7963 | 0.9647 | 0.9243 | 0.6536 |
| 16 | 0.9436 | 0.8867 | 0.7563 | 0.9028 | 0.8082 | 0.6517 | 0.9408 | 0.8823 | 0.7568 | 0.9397 | 0.8802 | 0.6183 |
| 17 | 0.9030 | 0.8106 | 0.7124 | 0.9142 | 0.8250 | 0.6630 | 0.9168 | 0.8344 | 0.7166 | 0.9190 | 0.8390 | 0.6023 |
| 18 | 0.8383 | 0.7197 | 0.6133 | 0.6630 | 0.7230 | 0.6072 | 0.8832 | 0.7816 | 0.6727 | 0.8824 | 0.7818 | 0.5384 |
| 19 | 0.9033 | 0.7825 | 0.6770 | 0.9098 | 0.7941 | 0.6581 | 0.9216 | 0.8273 | 0.7235 | 0.9150 | 0.8102 | 0.5802 |
| 20 | 0.8873 | 0.8085 | 0.6836 | 0.8964 | 0.8200 | 0.6949 | 0.9124 | 0.8507 | 0.7695 | 0.9048 | 0.8380 | 0.5988 |
| **Mean ± Std.** | **0.9197 ± 0.0315** | **0.8377 ± 0.0614** | **0.7306 ± 0.0531** | **0.9016 ± 0.0600** | **0.8188 ± 0.0452** | **0.6755 ± 0.0399** | **0.9267 ± 0.0250** | **0.8494 ± 0.0569** | **0.7417 ± 0.0509** | **0.9265 ± 0.0240** | **0.8496 ± 0.0515** | **0.6066 ± 0.0337** |

# 5. Discussion

Based on the data shown in Table 2, and the statistical analysis performed with the Wilcoxon test, it can be concluded that using two signals improves the results over using a single signal as an input, whereas the ensemble model shows no advantage with respect to the double-signal-input model. The main reason why the ensemble was unable to improve the results of the double-signal input lays on the incapacity of the single-input-signal architectures to identify information that was not previously captured by the double-signal model.

As previously mentioned in section 3.1, the expanded Sleep-EDF dataset is the most commonly used dataset in the literature and this was our main reason to choose it. Table 3 contains a comparison of the results of previously published works over the same dataset. It should be noted that a proper comparison may be very difficult due to the diversity of splitting strategies used in these works.

Table 3: Comparison of accuracy, Cohen's Kappa and F-1 score of the works that performed their test on the Sleep-EDF

| Reference | Technique | Split type | CV | Accuracy | Kappa | F1-macro |
|---|---|---|---|---|---|---|
| (Tsinalis et al. 2016) | CNN | Record | 20-fold CV | 0.75 | 0.65 | 0.70 |
| (Supratak et al. 2017) | CNN-LSTM | Record | 20-fold CV | 0.82 | 0.76 | 0.77 |
| (Zhu et al. 2014) | Difference Visibility graph, SVM | Example | 10-fold CV | 0.89 | 0.79 | 0.73 |
| (Hassan and Bhuiyan 2016a) | EMD domain, ensamble | Example | 0.6/0.05/0.35 | 0.87 | 0.82 | 0.80 |
| (Hassan and Bhuiyan 2016b) | EMD, bootstrap aggregation | Example | 0.5/0.5 | 0.89 | 0.85 | 0.83 |
| (Hassan and Bhuiyan 2016c) | Wavelet transform, Spectral features, random forest | Example | 0.5/0.5, 20-fold average | 0.88 | 0.84 | 0.80 |
| (Hassan and Bhuiyan 2017) | EMD, random undersampling boosting | Example | 0.5/0.5, 20-fold average | 0.83 | 0.76 | 0.74 |
| **This work** | **CNN (Fpz-Cz)** | **Record** | **20-fold CV** | **0.9197** | **0.8378** | **0.7307** |
| | **CNN (Pz-Oz)** | **Record** | **20-fold CV** | **0.9110** | **0.8188** | **0.6755** |
| | **CNN (Fpz-Cz & Pz-Oz)** | **Record** | **20-fold CV** | **0.9266** | **0.8594** | **0.7417** |
| | **Ensemble** | **Record** | **20-fold CV** | **0.9265** | **0.8496** | **0.6066** |

According to the splitting strategy, the research studies can be divided into two main groups. First, some authors performed what is called an 'Example' split. In this case, data from the same patient can be present in training and testing sets. On the other hand, the 'Record' splitting strategy ensured that the data from a patient was used only in training or testing sets, which is also closer to a real application of the system. From the authors' point of view, the preferred option is to make the split according to the patient because otherwise the independency of the test could be contaminated by having information from the same source in the training set. Table 3 provides details on the splitting strategy followed in each work. The split column specifies "Record" when the split was made by patient record, or "Example", if the split was made based only on the signal sections, without considering the patient dependency.

Table 4: Comparison of the results with the other Deep Learning published approaches

| Reference | Dataset | Channels | Patients | CV | Input size | Accuracy | Kappa | F1-macro | Trainable parameters |
|---|---|---|---|---|---|---|---|---|---|
| (Tsinalis et al. 2016) | Sleep-EDF | Fpz-Cz | 20 | 20-fold | 3000 | 0.75 | 0.65 | 0.70 | 1,114,000 |
| (Supratak et al. 2017) | Sleep-EDF | Fpz-Cz | 20 | 20-fold | 3000 | 0.82 | 0.76 | 0.77 | 546,525,189 |
| | Sleep-EDF | Pz-Cz | 20 | 20-fold | 3000 | 0.798 | 0.72 | 0.731 | |
| | MASS | F4-EOG | 31 | 31-fold | 3840 | 0.862 | 0.817 | 0.80 | |
| (Sors et al. 2018) | SHHS-1 | C4-A1 | 5728 | 0.5/0.2/0.3 | 15000 | 0.87 | 0.81 | 0.78 | 199,068,478 |
| **This work** | **Sleep-EDF** | **Fpz-Cz** | **20** | **20-fold** | **3000** | **0.9197** | **0.8378** | **0.7307** | **13,485** |
| | **Sleep-EDF** | **Pz-Oz** | **20** | **20-fold** | **3000** | **0.9110** | **0.8188** | **0.6755** | **13,485** |
| | **Sleep-EDF** | **Fpz-Cz Pz-Oz** | **20** | **20-fold** | **6000** | **0.9266** | **0.8594** | **0.7417** | **13,625** |
| | **Sleep-EDF** | **Ensamble** | **20** | **20-fold** | **6000** | **0.9265** | **0.8496** | **0.6066** | **40,595** |

Moreover, among the works which performed the split by example, there are many strategies which were also specified in the table. CV column shows the type of split performed which may be X-fold when some cross validation was used or a collection of three values representing percentages for training, validation and testing size when no cross validation was used.

Finally, the obtained accuracy, Cohen's Kappa and F1-score are also shown as reference values in Table 3. An analysis of the results clearly shows an improvement compared to previous works on the same dataset. Even though in the cases where the authors performed an Example splitting, which is supposed to be easier, the proposed architecture achieved better results in accuracy and Cohen's Kappa. In fact, if the focus is on the Deep Learning works, the proposed model outperforms the results in accuracy, Kappa and F1. Comparing the Cohen's Kappa values with those obtained by (Tsinalis et al. 2016) and (Supratak et al. 2017) our results are better aligned with the results provided by the human experts than the two aforementioned works. Moreover, the agreement of this model with the human experts is nearly the same as the interobserver agreement ratio measured in (Norman et al. 2000).

With the remaining works, the comparison is more difficult due to the previously discussed dataset splitting strategy. However, this work shows better results in accuracy and Kappa value than any other previously published work dealing with the same dataset.

Eventually, there are not many works to compare to, if the focus is exclusively on Deep Learning approaches. In order to increase the possibilities, some recently published works were included in Table 4 even though some of them were developed on different datasets and with different input channels. All these works used a record split on the datasets and the objective was to rate according to the AASM guidelines. Although the comparison is very difficult, there are two points which should be taken into account. First, the proposed solution is the smallest among the published approaches and, second, the accuracy and kappa values are better than those obtained in the remaining works. As already mentioned, if only the works dealing with the same dataset and the same splitting strategy were considered, the results obtained by this architecture would be clearly better and the number of parameters to train would be significantly smaller.

# 6. Conclusions

From the results contained in section 5, three main conclusions can be drawn. First and foremost, tests have shown the advantages of using at least two signals as inputs instead of only one, as it has been the case in the literature so far. In fact, using the same architecture with 3 different inputs, Fpz-Cz, Pz-Oz or both signals at the same time, proved that the results obtained by the network with two inputs were statistically better than the ones obtained with a single signal.

Moreover, tests have also shown that there was no difference between the double-signal-input model and the ensemble model, which combined the three best architectures in a single meta-classifier. This fact points out to the impossibility of the single-signal-input models to identify information which the double-signal-input model is not able to. Therefore, when a PSG has to be labelled, it seems convenient to include as many information in the input as possible in order to help the network extract the valuable relationships among the different available data.

Second, tests have also shown that it was possible to develop smaller networks which improved the results. Increasing the pressure on the features extracted by the convolutional layers seems profitable, which is the opposite of what some common approaches in the 2D and 3D signals did. When 1D signals are processed, tests contained in Table 4 show an improvement when less features are extracted from each layer. The reduction in the number of features caused by the pooling layers have a positive outcome, which may be due to an improvement in the meanness of the features.
Finally, it should be pointed out that the proposed architecture improves the previous state-of-the-art and performs better in the presented 5-class problem. In fact, regardless of the input, the architecture improves the results of the previously published works in terms of Accuracy and Kappa values.

# 7. Future Work

The results of this work open several research lines. First, further experiments are needed, including different channels and using other datasets. It would be also interesting to include different signals from a PSG in the input to identify whether the network is able to find patterns not only in the EEGs, but also when using electrooculograms, electrocardiograms or electromyograms in the same architecture.

Therefore, having access to other datasets, such as the aforementioned MASS and SHHS-1, would be necessary.

Secondly, it should be noted that, while in other knowledge areas Deep Learning has had a deep impact, the 1D signal processing still has very few models. This fact limits the applicability of some of the most modern approaches such as fine-tuning (Yosinski et al. 2014), generative adversarial networks (Goodfellow et al. 2014) or distillery (Goodfellow et al. 2016). Therefore, a research line in the foreseeable future should be increasing the number of the available models to apply the aforementioned new approaches.

Focusing exclusively on the Deep Learning, it may be interesting to conduct experiments which evaluate the influence on the performance of the pressure in the extracted features. By developing models with approaches for further reduction of parameters such as the depth wise convolution layers (Chollet 2017), very valuable information could be obtained which could lead to modifications when a network is being designed. Nowadays, since many researchers increase the number of features extracted by the convolutional kernel without a clear reason, identifying methods or mechanisms to limit that number could be the key for more rational developments.

Finally, the data from patients under treatment with drugs have been excluded from the tests in this paper. It would be interesting to explore whether there are differences between the EEGs of healthy patients and those using medication. Combining both points of view, the next step could be the development of a Siamese neural network (Ranjan et al. 2017) which would be able to solve both problems at the same time, labelling the section and identifying whether the data come from a patient using drugs or not. This approach has been used in images showing an improvement when related problems are solved at the same time.

# Acknowledgement

The authors would like to thank the support from NVidia corp., which granted the GPU used in this work. They also acknowledge the support from the CESGA, where many of the preliminary tests were run. This work is supported by the project granted by the Carlos III Health Institute (PI17/01826) within the Spanish National plan for Scientific and Technical Research and Innovation 2013–2016 and the European Regional Development Funds (FEDER). The authors would also like to acknowledge the support from the Galician government in the form of grants (ED431D 2017/23, ED431D 2017/16, ED431G/01) and that from the European Union in the form of ERDF funds.

# Compliance with Ethical Standards

This work was supported by NVidia corp., which granted the GPU used in this work.
Moreover, it was also partially supported by the project granted by the Carlos III Health Institute (PI17/01826) within the Spanish National plan for Scientific and Technical Research and Innovation 2013–2016 and the European Regional Development Funds (FEDER).
Besides the affiliation of the researchers, there is no other conflict of interest.